\documentclass[conference]{IEEEtran}
\IEEEoverridecommandlockouts
\usepackage{cite}
\usepackage{amsmath,amssymb,amsfonts}
\usepackage{algorithmic}
\usepackage{graphicx}
\usepackage{subfig}
\usepackage{multirow}
\usepackage{textcomp}
\usepackage{xcolor}
\def\BibTeX{{\rm B\kern-.05em{\sc i\kern-.025em b}\kern-.08em
    T\kern-.1667em\lower.7ex\hbox{E}\kern-.125emX}}
\begin{document}

\title{A Binary Classification Social Network Dataset for Graph Machine Learning}

\author{\IEEEauthorblockN{Adnan Ali*, Jinlong Li, Huanhuan Chen, AlMotasem Bellah Al Ajlouni}
\IEEEauthorblockA{\textit{School of Computer Science and Technology} \\
\textit{University of Science and Technology of China}\\
Hefei, China \\
*adnanali@mail.ustc.edu.cn, (jlli, hchen@ustc.edu.cn), motasem@mail.ustc.edu.cn }
}

\maketitle

\begin{abstract}
Social networks have a vast range of applications with graphs. 
The available benchmark datasets are citation, co-occurrence, e-commerce networks, etc, with classes ranging from 3 to 15. However, there is no benchmark classification social network dataset for graph machine learning. 
This paper fills the gap and  presents the \textbf{Bi}nary Classification \textbf{S}ocial \textbf{N}etwork \textbf{D}ataset (\textit{BiSND}), designed for graph machine learning applications to predict binary classes. 
We present the BiSND in \textit{tabular and graph} formats to verify its robustness across classical and advanced machine learning. 
We employ a diverse set of classifiers, including four traditional machine learning algorithms (Decision Trees, K-Nearest Neighbour, Random Forest, XGBoost), one Deep Neural Network (multi-layer perceptrons), one Graph Neural Network (Graph Convolutional Network), and three state-of-the-art Graph Contrastive Learning methods (BGRL, GRACE, DAENS). 
Our findings reveal that BiSND is suitable for classification tasks, with F1-scores ranging from 67.66 to 70.15, indicating promising avenues for future enhancements. 

\end{abstract}

\begin{IEEEkeywords}
Graph machine learning, binary classification, machine learning, graph neural networks, graph contrastive learning
\end{IEEEkeywords}

\section{Introduction}
A graph is a data structure employed to model the relationship between objects where objects are called nodes/vertices and connections between nodes are called edges \cite{NegAmplify}.  
Graphs can be directed or undirected, where directed graphs contain directional edges indicating a one-way relationship, and undirected graphs depict a two-way relation. 
Graphs have been vastly utilized in multiple computer science domains, including citation networks \cite{CiteseerDataset, PubMed2}, e-commerce networks \cite{AmazonDatasets}, co-occurrence networks \cite{ActorDataset}, transportation system networks, recommender systems \cite{10152806}, social networks \cite{AdnanAliTemp} are few of many. 
However, the nature of the graph elements can differ in each domain. 
In a citation network, words are presented as nodes, and edges are created based on the  usage of the words in different papers \cite{CoraDataset}. In contrast, in e-commerce, products are nodes linked by frequently bought products together \cite{AmazonDatasets}. In co-occurrence networks, elements (e.g., actors) are presented as nodes, while edges indicate their co-occurrence in the movie \cite{ActorDataset}. 
\par
Similarly, social networks are one of the most prominent applications of graphs, where graphs are implemented for various purposes, including classifying nodes\cite{NegAmplify}, clustering the related nodes \cite{liu2023hard}, detecting temporal patterns \cite{AdnanAliTemp}, detect influencers \cite{ZHANG2021131}, etc.  
Social networks refer to the platform to connect individuals in cyberspace. 
Famous social network platforms include Facebook, Instagram, X (Twitter), YouTube, WhatsApp, TikTok, and WeChat, where each of these platforms has millions of active users every day \cite{AdnanAliTemp}.
Graphs are the backbone of these platforms, where online social networks and graphs are intricately linked by representing relationships and interactions among users \cite{SpammerDataset}. 
Graph theory provides a robust framework for modeling these networks, where users are represented as nodes (or vertices), and their connections (friendships, follows, interactions) are represented as edges.
\par
Presenting social network users and their interactions on graphs enables the ability to apply graph theory and graph machine learning algorithms on graphs to analyze the various properties of social networks, such as connectivity, centrality, user classification \cite{NegAmplify}, and clustering \cite{liu2023hard}. 
For instance, centrality measures can identify influential users within the social network, which is crucial for marketing and information dissemination strategies \cite{ZHANG2021131}.  
Graph algorithms can be employed to detect communities within social networks. 
These communities represent groups of users more densely connected than the rest of the network \cite{comunityGuo2023}. 
User classification is performed on social networks to classify the users as spam or not \cite{SpammerDataset}. 
Clustering \cite{liu2023hard} is used to group similar properties users while the properties can be posted content, posting behavior, or metadata information. 
However, typical graph algorithms are limited to performing all these actions.
Recently, the machine learning community has been paying significant attention to graphs, and as a result, graph machine learning (GML) has come into existence. 
\par  
Graph Machine Learning is a machine learning subfield that analyzes and interprets graph data structure. \cite{CostaZhang_2022}. 
This approach is beneficial for tasks where relationships and interactions are as meaningful as the features of the individual entities \cite{thakoor2021bootstrapped}.
However, graphs are inherently complex compared to tabular, text, and visual data due to multiplex node relations.  
Graph Neural Networks (GNNs) are famous contributors to GML, which extend neural network architectures to operate on graph structures, allowing for learning both node and graph-level representations \cite{9770382}. 
GNNs are applied for representation learning, classification, clustering, embedding, and link prediction. 
\par
As previously mentioned, graphs have applications in various domains where social media is one of the most significant due to its vastness and applications, including user behavior, community detection, and link prediction. 
However, The available benchmark datasets used by various studies \cite{DeepGrace2020,thakoor2021bootstrapped,CostaZhang_2022,FebAA.2207.01792,10025823} are citation networks (Cora, CiteSeer, DBLP, PubMed, WikiCS, ), co-occurance networks (Actor, Co-Author CS, and Physics), and e-commerce networks (Amazon Photo and Computers) offers limited variations. 
Regardless of the diversity and vastness of social media, no single social network graph dataset is available for graph machine learning in PyTorch geometric datasets \cite{FeyLenssen2019} and is not used in SOTA methods. 
\par 
The available benchmark datasets are presented in Table \ref{app:table:datastatsProperties}, and their detailed description is given in the next section (Section \ref{app:Datasets}). 
The properties of available benchmark datasets (Table \ref{app:table:datastatsProperties}) indicate that benchmark datasets offer a wide variety of nodes, features, node degree, edge ratio, degree, and features. 
Furthermore, the node-to-edge ratio in benchmark datasets is high, indicating an average of many edges against one node. 
Regardless of the non-availability of social network datasets, no binary classification dataset is available in benchmark datasets. 
We identify two gaps: 1- the social networks dataset is unavailable, and 2: no binary classification dataset.
In this study, we aim to fill these two gaps, where contributions of this study are:
\begin{itemize}
    \item This study introduces a novel real-world binary classification dataset for social networks, termed the \textbf{Bi}nary Classification \textbf{S}ocial \textbf{N}etwork \textbf{D}ataset (BiSND). This dataset is designed explicitly for graph machine learning applications to ascertain the presence of users on X (Twitter). The dataset is provided in tabular and graph formats, with the graph structure available in three variations: node-only, undirected, and directed graphs.
    \item We assess the robustness of the dataset using various machine learning methodologies, including ensemble machine learning techniques, multi-layer perceptrons, graph neural networks, and self-supervised graph contrastive learning. The F1-scores across these methods range from 67 to 71, suggesting that the dataset is sufficiently robust for classification and indicating the potential for further enhancement in classification outcomes in future research. 
    \item Our investigation also aims to address the following two research questions based on the BiSND: Which format—tabular or graph datasets—proves to be more effective for downstream classification tasks? Which machine learning paradigm is superior for sparse graphs: supervised or self-supervised learning? 
\end{itemize}

The rest of the paper is formatted as follows: Section \ref{app:Datasets} analyzes the benchmark datasets and their application in various SOTA graph machine learning methods. 
Section \ref{sec:methodology} presents the methodology to create BiSND and outputs tabular and graph datasets. Section \ref{sec:spam:chap3results} verifies the robustness of BiSND with multiple methods and presents the results. The last section (Section \ref{sec:Conclusion}) concludes the findings of this paper.


\section{Benchmark Public Datasets}
\label{app:Datasets}
This section briefly introduces the node classification benchmark datasets used by graph machine learning state-of-the-art (SOTA) methods. 
The statistics of each dataset are presented in Table \ref{app:table:datastatsProperties}. 

\subsection{Cora}
Cora is a citation network dataset, introduced in \cite{CoraDataset} and is the most used dataset in graph machine learning research \cite{velikovi2019deep,LG2AR}.
Cora comprises 2,708 scientific publications (nodes) and 10,556 edges. Nodes are sparse binary word vectors that describe the presence of 1,433 unique words extracted from the document titles and abstracts. 
Each publication belongs to one of seven computer science research areas, forming the seven dataset classes. 
Cora is notable for its utility in self-supervised node classification tasks, allowing models to learn effective node representations. 
Dataset size and complexity are crucial when extrapolating model performance \cite{CostaZhang_2022}. 
However, Cora is a relatively modest size and limited complexity dataset, which might not fully represent the challenges of larger, more intricate graph datasets. 
Notable models using the Cora dataset for their performance evaluation are BGRL \cite{thakoor2021bootstrapped}, GRACE \cite{DeepGrace2020}, DGI \cite{velikovi2019deep}, MVGRL\cite{pmlr-v119-hassani20a}, LG2AR \cite{LG2AR}, COSTA \cite{CostaZhang_2022}, AF-GCL \cite{AFGCL}, and GRAM \cite{10025823} etc.


\subsection{CiteSeer}
The CiteSeer dataset is commonly used in graph machine learning and graph contrastive learning, introduced in a paper \cite{CiteseerDataset} and created by \cite{PubMed2}. 
The CiteSeer dataset consists of 3,327 scientific publications (nodes) from six fields of study (classes) and 9,104 edges representing the citation relationships between works. 
The CiteSeer's feature matrix consists of bag-of-words representations of documents, where each word is considered a feature and contains 3,703 unique words. 
The dataset's structure and size make it a valuable resource for evaluating graph contrastive learning algorithms \cite{}. 
However, similar to the Cora dataset, CiteSeer's relatively small scale, while dataset complexity, should be considered when generalizing model performance to more complex real-world graph data.
Notable self-supervised learning methods utilizing CiteSeer includes MVGRL \cite{pmlr-v119-hassani20a}, COSTA \cite{CostaZhang_2022}, GRAM \cite{10025823}, and DAENS \cite{DAENS}. 

\subsection{PubMed}
The PubMed \cite{PubMed2} dataset is a widely recognized benchmark in the graph machine learning community for node classification tasks. 
Like Cora and CiteSeer, it is a citation network of 19,717 scientific publications (nodes) interconnected through 88,648 citation links (edges), and each publication in the dataset is labeled with one of three classes. 
The feature vector for each node is a Term Frequency-Inverse Document Frequency (TF-IDF) weighted word vector of 500 unique words. 
The PubMed dataset's larger size and use of TF-IDF features distinguish it from smaller, binary word vector-based datasets like Cora and CiteSeer, offering a more challenging and realistic scenario for evaluating the performance of graph-based learning models.
PubMed is utilized in various SOTA methods to verify their performance, including GRACE \cite{DeepGrace2020}, DGI \cite{velikovi2019deep}, BGRL \cite{thakoor2021bootstrapped}, MVGRL\cite{pmlr-v119-hassani20a}, LG2AR \cite{LG2AR}, and GRAM \cite{10025823}. 

\subsection{Dblp}

The DBLP \cite{dblpdataset} dataset contains 17,716 nodes (papers) and 105,734 edges (citations), making it a sizable network for graph learning tasks. 
Each node is labeled with one of four classes based on the paper's topic, and the node features are derived from the paper's content, typically represented as sparse bag-of-words vectors.
Its scale and the specificity of citation relationships offer unique challenges for learning algorithms, such as handling large-scale data and capturing the nuances of citation patterns.
It is not widely utilized by SOTA methods, and we think the reason is its hard to improve accuracy. 
It is not utilized by MVGRL\cite{pmlr-v119-hassani20a}, LG2AR \cite{LG2AR},AF-GCL \cite{AFGCL}, and AdaS \cite{10181235}. 
Although, these methods have used CORA, CiteSeer, and PubMed datasets. 
The SOTA methods which verify their performance on DBLP, include DGI \cite{velikovi2019deep}, GRACE \cite{DeepGrace2020}, BGRL \cite{thakoor2021bootstrapped}, COSTA \cite{CostaZhang_2022}, FebAA \cite{FebAA.2207.01792} and GRAM \cite{10025823}.

\begin{table*}[t]
    \caption{Benchmark dataset properties and statistics. ND: Node Degree, IN: Isolated Nodes, SL= self-Loops, UD=Undirected,  BF: Binary Features  }
\label{app:table:datastatsProperties}
\begin{center}
\begin{tabular}{lllllllllllll} 
		\noalign{\smallskip}\hline\noalign{\smallskip}	
				
		Dataset & Nodes & Feat. &Edges& Class& ND & ND Ratio & N/E & Non 0s & IN & SL & UD & BF \\
		\hline\noalign{\smallskip}
				
		Cora&2708&1433&10556&7& 3.90 & 17/1 & 25.7\%&1.27\% & F & F & T & T\\
		CiteSeer&3327&3703&9104&6& 2.74 & 13/0.7 &36.5\%&0.85\%& T & F & T & T \\
		DBLP&17,716&1639&105734&4 &5.97&34/1&16.8\%&0.32\%& F & F & T & T\\
		PubMed&19717&500&88648&3& 4.50 & 29/1 &22.2\% & 10.0\%& F &F & T & F \\
		Actor&7600&932&30019&5 &3.95 & 19/0 &25.3\%&0.58\%& F & T & F & T \\
				
		\noalign{\smallskip}\hline\noalign{\smallskip}

		WikiCS&11701&300&431726&10 & 36.90 &229/0.4&2.7\%&99.99\%&T & T  & T &F \\
		Am.Comp&13752&767&491722&10 & 35.76 & 217/0.7 &2.8\%&34.84\%& T & F & T & T \\
		Am.Photo&7650&745&238162&8 & 31.13 & 168/0.95 &3.2\%&34.74\%& T & F & T & T \\
		Co.CS&18333&6805&163788&15 & 8.93 & 38/1 &11.2\%&0.88\%& F & F & T & T \\
				
		Co.Phy&34493&8415&495924&5 & 14.38 &64/1.5 &6.96\%&0.39\%& F & F & T & T \\
		\noalign{\smallskip}\hline\noalign{\smallskip}	
            BiSND&12788&19&430&2 & 0.01 &0.67/0 &0.30\%&80.13\%& F & T & F & F \\
		\noalign{\smallskip}\hline\noalign{\smallskip}	
\end{tabular}
\end{center}
\end{table*}

\subsection{Actor}
A network of actors who appear together in a movie is introduced in GEOM-GCN \cite{ActorDataset}.
In the Actor dataset, nodes represent actors, and edges indicate co-occurrence on the same Wikipedia page, forming a co-occurrence network. The dataset encompasses 7,600 nodes connected by around 30,019 edges. 
Each actor is labeled with one of five classes, corresponding to the genre of movies they predominantly act in.
The node features in this dataset are extracted from keyword occurrences in the actors' Wikipedia pages and encoded as multi-dimensional feature vectors.
It is not a well-utilized method in previous work while recently being utilized in FebAA \cite{FebAA.2207.01792}, DAENS \cite{DAENS}, and NegAmplify \cite{NegAmplify}.

\subsection{Amazon} 
The Amazon \cite{AmazonDatasets}  Photo and Computers datasets are tailored for graph-based machine learning tasks, focusing on items from Amazon's product co-purchasing network. Both of these datasets are used by various recent SOTA methods including GCA \cite{GCADBLP-abs-2010-14945}, BGRL \cite{thakoor2021bootstrapped}, LG2AR \cite{LG2AR}, AFGRL\cite{AFGRL}, COSTA \cite{CostaZhang_2022}, AF-GCL \cite{AFGCL}, AdaS \cite{10181235}, GRAM \cite{10025823}, FebAA \cite{FebAA.2207.01792}, DAENS \cite{DAENS}, and NegAmplify \cite{NegAmplify}.

\subsubsection{Photo}
This dataset represents a network of products in the `Photo' category on Amazon. 
The nodes in this graph are individual products, and the edges represent that two products are frequently bought together.
The Amazon Photo dataset comprises 7,650 nodes (products) and 238,162 edges (co-purchasing relationships).
Each node (product) is characterized by feature vectors (745 features) derived from reviews and product information, encoding information like user ratings and text descriptions.

\subsubsection{Computers}
Similar in structure to the Photo dataset, the Computers dataset focuses on products in the `Computers' category of Amazon.
It contains  13,381 nodes (products) connected by  491,722 edges (co-purchasing relationships).
Node features (767) in this dataset are also derived from product reviews and information, encapsulating user interactions and preferences.

\subsection{Wiki-CS}
The WikiCS dataset \cite{WikiCSdataset}  is a citation network derived from Wikipedia articles in the field of computer science. 
It comprises 11,701 nodes (articles), 300 features, and 216,123 edges (citations). 
Each node features a high-dimensional vector based on TF-IDF encoding of the article's text. Classified into 10 computer science fields, the dataset is primarily used for node classification tasks and is continuously used in recent methods same as Amazon datasets.

\subsection{Co-Author}
Co-Author \cite{AmazonDatasets} datasets present the academic co-authorship networks and are vastly utilized in recent methods due to the unique size and density of the dataset. 
Both are utilized in GCA \cite{GCADBLP-abs-2010-14945}, BGRL \cite{thakoor2021bootstrapped}, LG2AR \cite{LG2AR}, AFGRL\cite{AFGRL}, COSTA \cite{CostaZhang_2022}, AF-GCL \cite{AFGCL}, AdaS \cite{10181235}, GRAM \cite{10025823}, FebAA \cite{FebAA.2207.01792}, DAENS \cite{DAENS}, and NegAmplify \cite{NegAmplify}.
Co-Author Physics is a larger dataset compared to Co-Author CS. 

\subsubsection{CS}
The Coauthor CS dataset is tailored for graph machine learning, focusing on the co-authorship network in the field of Computer Science. 
It represents a dense network where nodes correspond to authors, and edges signify collaborative relationships based on co-authorship of scientific papers. 
The dataset contains 18,333 nodes (authors), 163,788 edges (co-authorship links), and 6805 features, making it a sizable network for graph analysis. 
The feature vector for each author is derived from keyword occurrences in their publications, resulting in each node having a high-dimensional feature vector. 

\subsubsection{Physics}
In parallel, the Coauthor Physics dataset mirrors the structure of the Coauthor CS dataset but is set within the field of Physics. 
This dataset includes 34,493 nodes (authors), 8415 features, and 495,924 edges (co-authorship links). 
Each node features a high-dimensional vector representing keywords from the author's publications. 
\begin{figure*}[t]
	\includegraphics[width=1\textwidth]{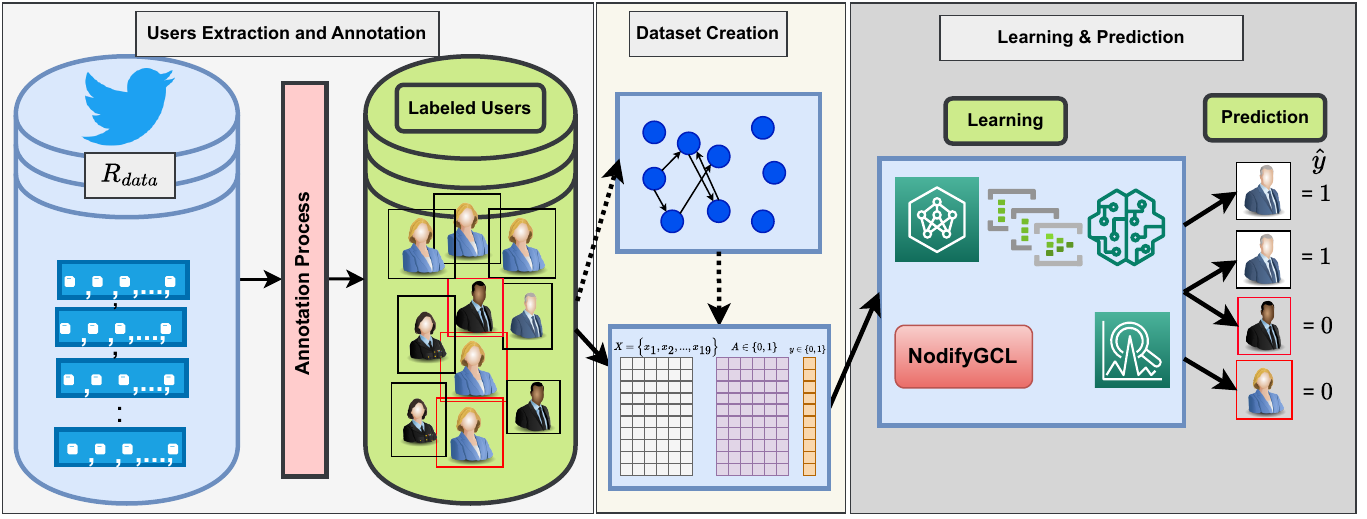}
	\caption{Model diagram to create and verify BiSND. First, Users are extracted and labeled, and then tabular and graph data is created from users. At last, supervised and self-supervised learning machine learning models are applied to check the authenticity of BiSND. NodifyGCL refers to node classification graph contrastive learning methods. }
	\label{fig:Spammermodel}       
\end{figure*}
\par
In summary, Table \ref{app:table:datastatsProperties} provides a comprehensive overview of ten benchmark datasets, encompassing academic, social, and commercial networks, including notable sources like Cora, CiteSeer, and DBLP, along with recent benchmark Amazon categories like Computers and Photos. 
It offers a detailed look into the structural intricacies of each dataset, covering aspects such as node and edge counts, class distribution, node connectivity, and feature types. 
However, the last row of Table \ref{app:table:datastatsProperties} presents the contribution of this study as \textit{BiSND} (\textbf{Bi}nary Classification \textbf{S}ocial \textbf{N}etwork \textbf{D}ataset) for graph machine learning.
In Table \ref{app:table:datastatsProperties}, it is just presented for comparison but the difference is explained in later sections. 
The next section presents the methodology for creating BiSND.

\section{Methodology}
\label{sec:methodology}
This section presents the \textbf{Bi}nary Classification \textbf{S}ocial \textbf{N}etwork \textbf{D}ataset (BiSND) and mentions the machine learning methods used to verify the robustness of BiSND. 
Three modules of this section are: 
\textit{Users Extraction and Annotation}, \textit{Dataset Creation}, and \textit{Learning \& Prediction}.

\subsection{Users Extraction and Annotation}
\label{sec:chap3:UserEx}
Figure \ref{fig:Spammermodel} presents the method to extract and annotate users where the annotation process is illustrated. 

\subsubsection{Dataset Downloading}
In our previous work \cite{AdnanAliTemp}, we extracted 1.1 million Twitter users' metadata data from Twitter, and  later  \cite{SpammerDataset} downloaded 190 thousand users, making a total of 1.29 million users. 
This dataset BiSND is the subset of the dataset presented in study \cite{SpammerDataset}. 
However, the previous version is only comprised of a tabular dataset. 
This study presents the subset of the tabular dataset and creates the graph dataset (Section \ref{sec:chap3GraphDataset}). 
We present the downloaded dataset of \cite{SpammerDataset} as $R_{data}$ in Figure \ref{fig:Spammermodel}.

\subsubsection{Annotation Process}
The $R_{data}$ users are authenticated from Twitter to verify their existence on Twitter and collected in a Set $U$. Formally, let $U=\{u_1,u_2,u_3,\ldots,u_n\}$ is the set of users and $U \subset R_{data}$.
If a user exists on Twitter, it is marked as 1, and if deleted, it is marked as 0.
Formally,  let $y$ be the label set where $y \in \{0,1\}^{N\times 1}$ and $0$ indicate that user $u_i$ is deleted from Twitter and $1$ indicates the user's existence. 
Set $U$ comprises two types of users, deleted $U_d$ and existing $U_e$. 
\begin{equation}
	U = U_d \cup U_e 
\end{equation}
Annotation is the process to label deleted users $U_d$ with $0$, presented as  $U_d=\{u \in U_d:f(u)=0\}$, and  existing users $U_e$ as $1$, presented as $U_e=\{u \in U_e:f(u)=1\}$, where $U_d \cap U_e = \phi$. 
The outcome of this section is users set $U$ and list of labels $y$, where $U= U_d \cup U_e$, and $y \in \{0,1\}^{N\times 1}$. 

\subsection{Tabular and Graph Dataset Creation}
\label{sec:spamuserDataset}
This section creates \textit{tabular} and \textit{graph} datasets using $U$ and $y$.
We create a feature matrix $X$ from extracted 19 metadata features of $U$, where $X \in \mathbb{R}^{N \times F}$ collects all users' features, $N$ is the number of users, and $F$ is the number of features. 
Users can be presented as $X_i$, while $X_i=U$ and $N=|X_i|=|U|$. 
In contrast, columns of $X$ can be presented as $X_j=\{x_1,x_2,x_3,\ldots,x_i\}$, where $F=|X_j|=19$ as 19 identified features.
Hence, $X_{ij}$ represents the $i$-th user's $j$-th feature.
Figure \ref{fig:Spammermodel} illustrates the feature matrix $X$. 
Column-wise, we normalize the features of $X$, depicting that the feature values range between 0 and 1. 
We create two types of datasets from them with metadata-based extracted features.

\subsubsection{Tabular Dataset}
$X$ is a feature matrix comprised of rows and columns, $Y$ is a set of labels, and we present the combined $X$ and $y$ as a tabular dataset. 
Figure \ref{fig:Spammermodel} illustrates creating the feature-based dataset with a solid arrow from \textit{Labeled Users} to \textit{Dataset Creation}. 

\subsubsection{Graph Dataset}
\label{sec:chap3GraphDataset}
A graph is comprised of nodes and edges. We present users as nodes, and if a user $u_i$ mentions the $u_j$ in the tweet, they will have an edge.
Figure \ref{fig:Spammermodel} illustrates that a graph-based dataset has three parameters, $X, A$ and $y$, where $X$ is the feature matrix, $A$ is the adjacency matrix, $y$ is a list of labels, and $G=(X, A)$ is the user graph. 
There are three variations of the graph dataset: 
\paragraph{Only Nodes}: Only users without any link, indicating no communication between users.  
\paragraph{Undirected Graph}: If a user $u_i$ mentions the $u_j$, they will have an undirected link, offering the unclearity of who communicated with whom. 
\paragraph{Directed Graph}: If a user $u_i$ mentions the $u_j$, there will be link from $u_i$ to $u_j$, indicating the one-way communication. 

\subsection{Learning \& Prediction}
\label{sec:chap3:classificaiton}
Learning is applying machine and deep learning to predict the user's class. 
We apply  four {Machine Learning} (ML), one {Deep Neural Networks} (DNN), one {Graph Neural Networks} (GNN), and three {Graph Contrastive Learning} (GCL) classifiers for learning and prediction to verify the robustness of our proposed BiSND, where these models aim to predict the class of user set $U$. 
It is essential to notice that ML, DNN, and GNN are \textit{supervised learning} while GCL is \textit{self-supervised learning}, where training is performed without labels. 
The further breakdown is as follows: the ML methods include Decision Trees, K-Nearest Neighbour, Random Forest, and XGBoost; DNN includes multi-layer perceptions; GNN includes GCN \cite{KipfW16} and GCL includes three SOTA methods: BGRL \cite{thakoor2021bootstrapped}, GRACE\cite{DeepGrace2020}, and DAENS \cite{DAENS}.

\section{Experimental Results}
\label{sec:spam:chap3results}
This section presents the \textit{Experimental Tools}, performs experiments on our proposed dataset \textbf{BiSND} to verify the robustness of BiSND using \textit{Machine Learning}, \textit{Deep Neural Networks}, \textit{Graph Neural Networks}, and \textit{Graph Contrastive Learning}. 

\subsection{Experimental Tools}
We use NetworkX \cite{SciPyProceedings11}, Scikit-learn, PyTorch, and PyTorch Geometric \cite{FeyLenssen2019} for dataset construction, machine learning, deep learning, and graph deep learning implementation, respectively. 
All experiments are an average of twenty executions. We measure the classification performance on metrics described in Section \ref{EvaluationMetrics}. 


\subsection{Accuracy Metrics}
\label{EvaluationMetrics}
BiSND is a binary classification dataset with two classes: True and False. Existing users are labeled True (1), and deleted users are labeled False (0) in BiSND. 
All of the ML learning algorithms are tested using the metrics given below. However, before explaining the metrics, we must understand the terms below. 
\begin{itemize}
\item \textbf{True Negative ($TN$)}
If the actual label is negative and the model predicts it as negative, it is called the true negative and presented as $TN$. 
\item \textbf{True Positive ($TP$)}
If the real label is true and the model predicts it as true, it is called the True positive and presented as $TP$. Ideally, the TP value should be higher
\item \textbf{False Negative ($FN$)}
False indicates a false prediction. If a value is negative, but the model predicts it as right, it is false negative and presented as $FN$. 

\item \textbf{False Positive ($FP$)}
It indicates that the model predicts positive. However, the actual label is negative. It is presented as $FP$. 
\end{itemize}

\subsubsection{Accuracy} Accuracy refers to classification accuracy as the ratio of correct predictions against total predictions.  
Per the above definition, $TP$ and $TN$ are true predictions. Hence, the equation of Accuracy is:

\begin{equation}
	Accuracy=\frac{TP+TN}{TP+TN+FP+FN}
\end{equation}


\subsubsection{Precision} 
It is the ratio between true positive predictions and total positive predictions presented as:
\begin{equation}
	Precision =\frac{TP}{TP+FP}
\end{equation}

\subsubsection{Recall} 
It is the ratio of true positives and the sum of true positives and false negatives presented as: 
\begin{equation}
	Recall=\frac{TP}{TP+FN}
\end{equation}

\subsubsection{F1-score:} It is  \textit{Harmonic Mean} between precision and recall, it is one of the most used classification metrics and better than accuracy and presented as:    
\begin{equation}
	\text{F1-score} = 2\times \frac{Precision \times Recall}{ Precision + Recall}
\end{equation}
In our experiment context, accuracy and F1-score micro produce equal results due to a balanced dataset. Hence, between accuracy and F1-score, \textit{we only consider the latter}.      

\subsubsection{Jaccard score:} 
It depicts how similar two sets (real labels and predicted labels) are and is calculated as follows: 
\begin{equation}
	J(y,\hat{y}) = \frac{|y \cap \hat{y}| }{ |y \cup \hat{y}|}
\end{equation}
where $y$ and $\hat{y}$ refers to true and predicted labels, respectively. 

\subsubsection{Time}
The sum of training and testing time. 

\subsection{Machine Learning Results}
Here, we present the results using the machine learning classifiers. 
All classifiers belong to the supervised learning paradigm, and we use the four classification techniques below to verify the accuracy of the BiSND dataset. 
Machine learning classifiers are: 1- Decision Tree (DT), 2- K-Nearest Neighbors (KNN), 3- Random Forest (RF), and 4- XGBoost (XGB).

\subsubsection{Decision trees}
Decision trees solve the problem by splitting the data based on features where decisions are at leaves.
We perform decision tree experiments with three criteria \textit{gini, entropy} and \textit{log loss} with tree depth 1 to 20. 
\par
\begin{figure}
	\centering
	\subfloat[F1-Score]{\label{DTR:DT_F1Score}\includegraphics[width=0.25\textwidth,height=4cm]{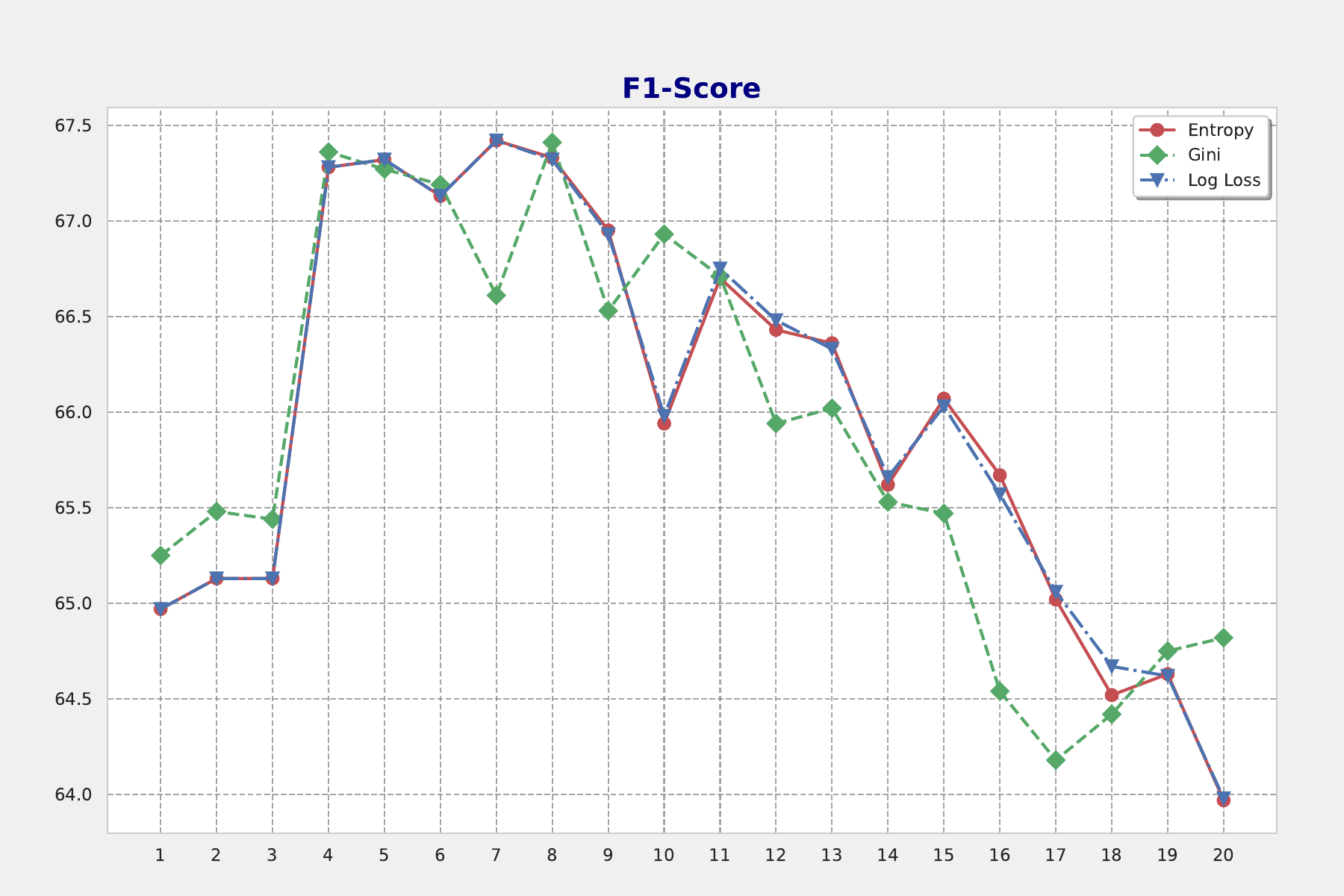}} 
	\subfloat[Precision]{\label{DTR:DT_Percision}\includegraphics[width=0.25\textwidth,height=4cm]{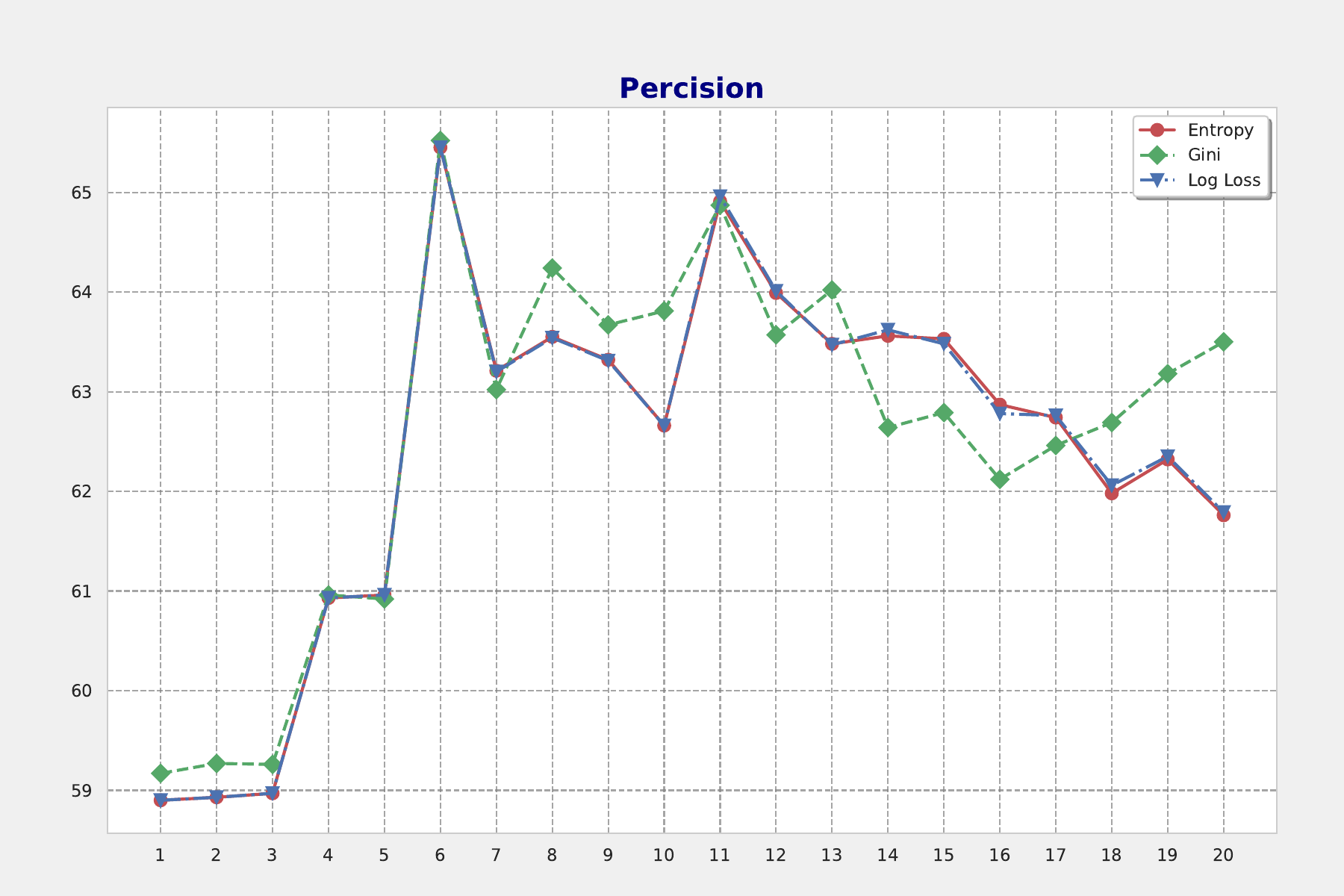}}\\
	\subfloat[Recall]{\label{DTR:DT_Recall}\includegraphics[width=0.25\textwidth,height=4cm]{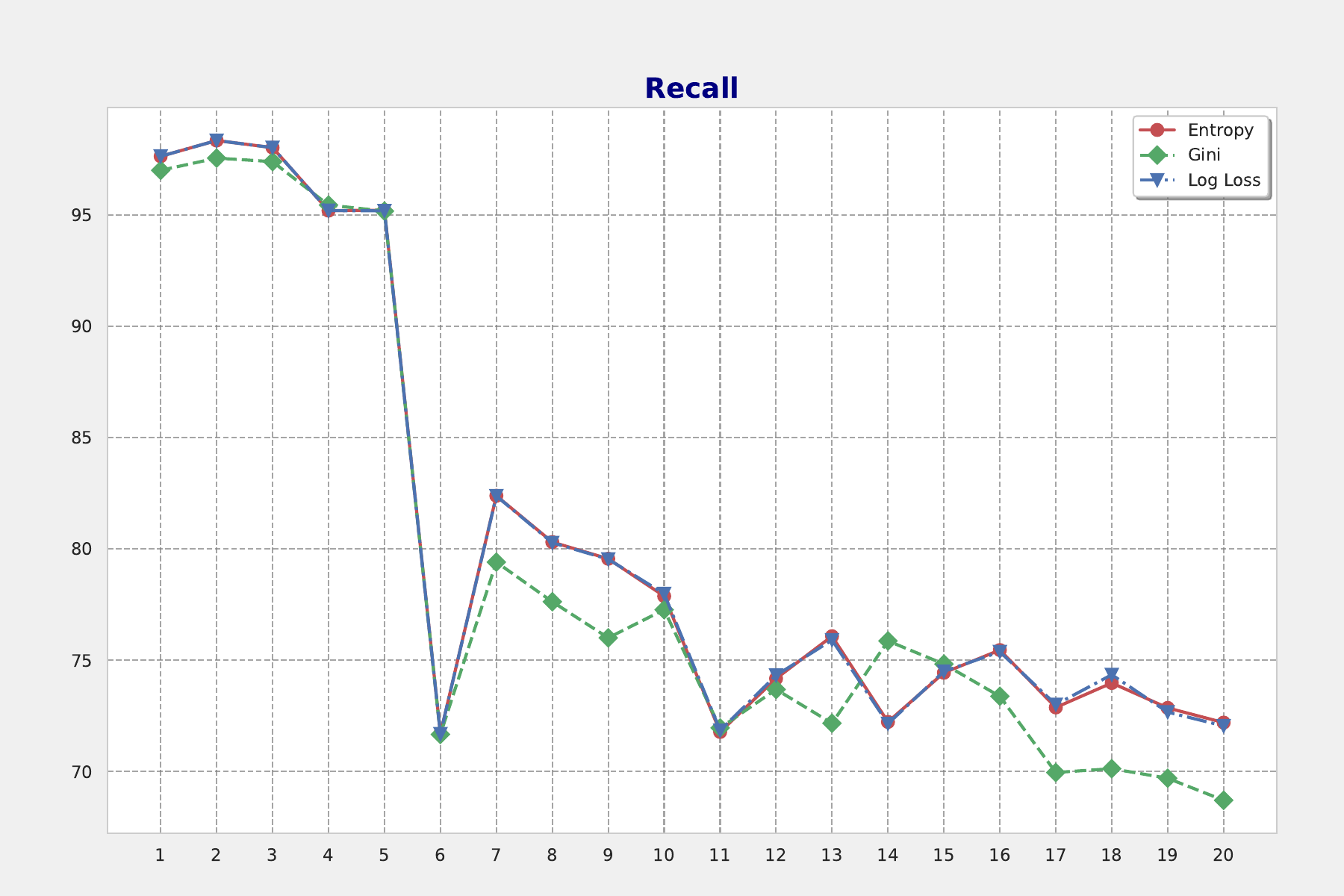}}
	\subfloat[Jaccard Score]{\label{DTR:DT_JScore}\includegraphics[width=0.25\textwidth,height=4cm]{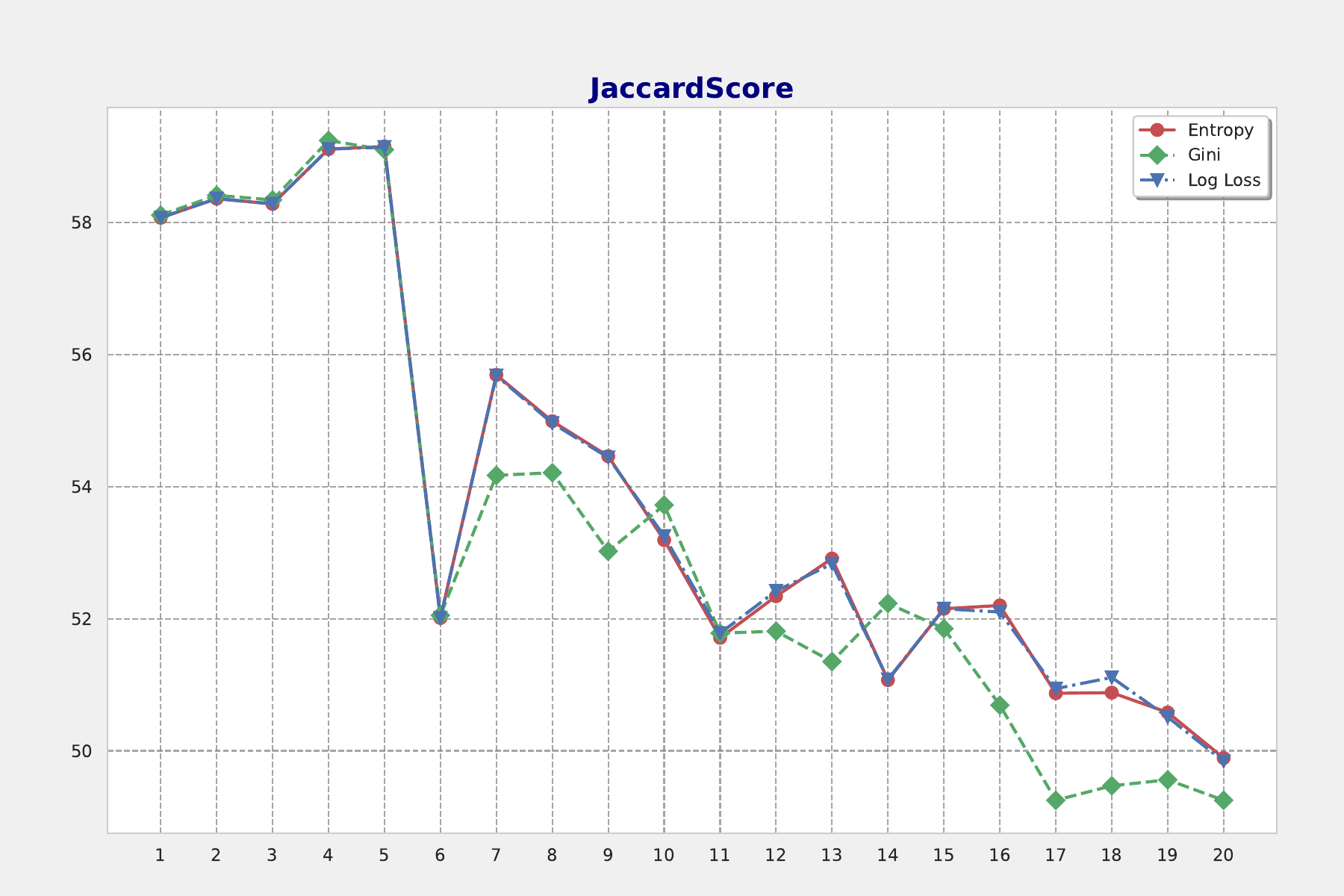}}\\
	\caption{Decision tree results on four metrics.} 
	\label{fig:Dt_results}
\end{figure}

Figure \ref{fig:Dt_results} presents the decision tree results on all metrics, and Figure \ref{DTR:DT_F1Score} shows the F1-Scores for three classification methods—Gini, Log Loss, and Entropy—across different depths. 
Initially, all three start similarly; however, from depths 1 to 5, Entropy and Gini surge ahead while Log Loss lags with a more moderate increase. 
In the mid-depths (5 to 10), Gini fluctuates but remains generally high, contrasting with Entropy and Log Loss, which  give identical results and gradually decrease. 
After a depth of 10, a notable decline in F1-Scores is seen for all methods, with Gini dropping more steeply.
A significant performance dip for Gini at depth 15 suggests a less effective model configuration. 
Towards the end, all methods slightly recover, with Log Loss showing a steadier performance and not dipping as significantly as the others. 
This pattern indicates that while Gini yielded higher initial F1 scores, its performance is less consistent, potentially due to overfitting or data variability. In contrast, Log Loss and Entropy provide more consistent and better performance, especially between 4 to 8 depth. Overall, Table \ref{tab:DTClas} indicates that Entropy gives the best F1-Score. 
 
\par
Figure \ref{DTR:DT_Percision} illustrates the precision metric of a decision tree model at various depths for three different split criteria: Entropy, Gini, and Log Loss. 
Precision measures the proportion of true positives in the positives predicted by the model. 
All three curves begin with low precision at shallow tree depths but experience a significant increase between depths 4 and 11. 
Following this rise, the precision values exhibit volatility, with several peaks and troughs indicating variability in model performance at different depths. 
Despite these fluctuations, the trend suggests that the precision generally stabilizes or slightly decreases as the tree depth increases. 
Entropy and Log loss show similar fluctuation patterns; in contrast, Gini has more fluctuations. 
Overall, the chart indicates that the precision of all methods is subject to change across different tree depths, and there is no consistent best performer throughout. However, as per Table \ref{tab:DTClas}, Log Loss has the best results across all tree depths and criteria. 
\par
Figure \ref{DTR:DT_Recall}  illustrates the recall value where the y-axis measures the proportion of actual positives correctly identified (true positives) out of all actual positives (true positives and false negatives). 
All three criteria show a steep decline after the first five depths, with  experiencing a sharp drop between depths 3 and 6 before stabilizing. 
Despite some fluctuations, the recall metric decreases as the depth increases. 
Figure \ref{DTR:DT_JScore} illustrates the Jaccard Score graph represents the similarity between the predicted and actual sets of positives, with higher scores indicating better model performance. 
Similar to the Recall graph, the Jaccard scores start higher but demonstrate a marked decrease as depth increases, particularly beyond depth 5. 
\par
Table \ref{tab:DTClas} concludes these experiments where all these results are an average of twenty executions. 
The table shows that entropy yields the best results in the F1-Score, Recall, and Jaccard Score.
However, Log Loss is the fastest and has better precision while maintaining a good F1-Score. 

\begin{table}[h]
	\centering
	\caption{Classification accuracy of decision tree algorithm where finest values are bold. DTC= Decision Tree Criterion, D= Depth of Tree, F1=F1-score, Pre= Precision, Rec= Recall, JS= Jaccard Score, and T=Time}
	\label{tab:DTClas}       
	\begin{tabular}{lllllll}
		\hline\noalign{\smallskip}
		DTC & D & F1 & Pre & Rec & JS &  T \\
		\hline\noalign{\smallskip}
		Entropy &  7 &  \textbf{67.42}   & \textbf{63.21}  & \textbf{82.39}  &\textbf{55.69} & 0.19\\
		Log Loss & 8 &   67.41   & 64.24  & 77.63  & 54.21 & \textbf{0.15}\\
		GINI &  6  & {67.42}   & 63.20  & {82.38}  &{55.68 }& {0.19}\\
		\noalign{\smallskip}\hline
	\end{tabular}
\end{table}

\subsubsection{Random Forest}
\begin{figure}[t]
\centering
\subfloat[F1-Score]{\label{RFC:RF_F1Score}\includegraphics[width=0.23\textwidth,height=4cm]{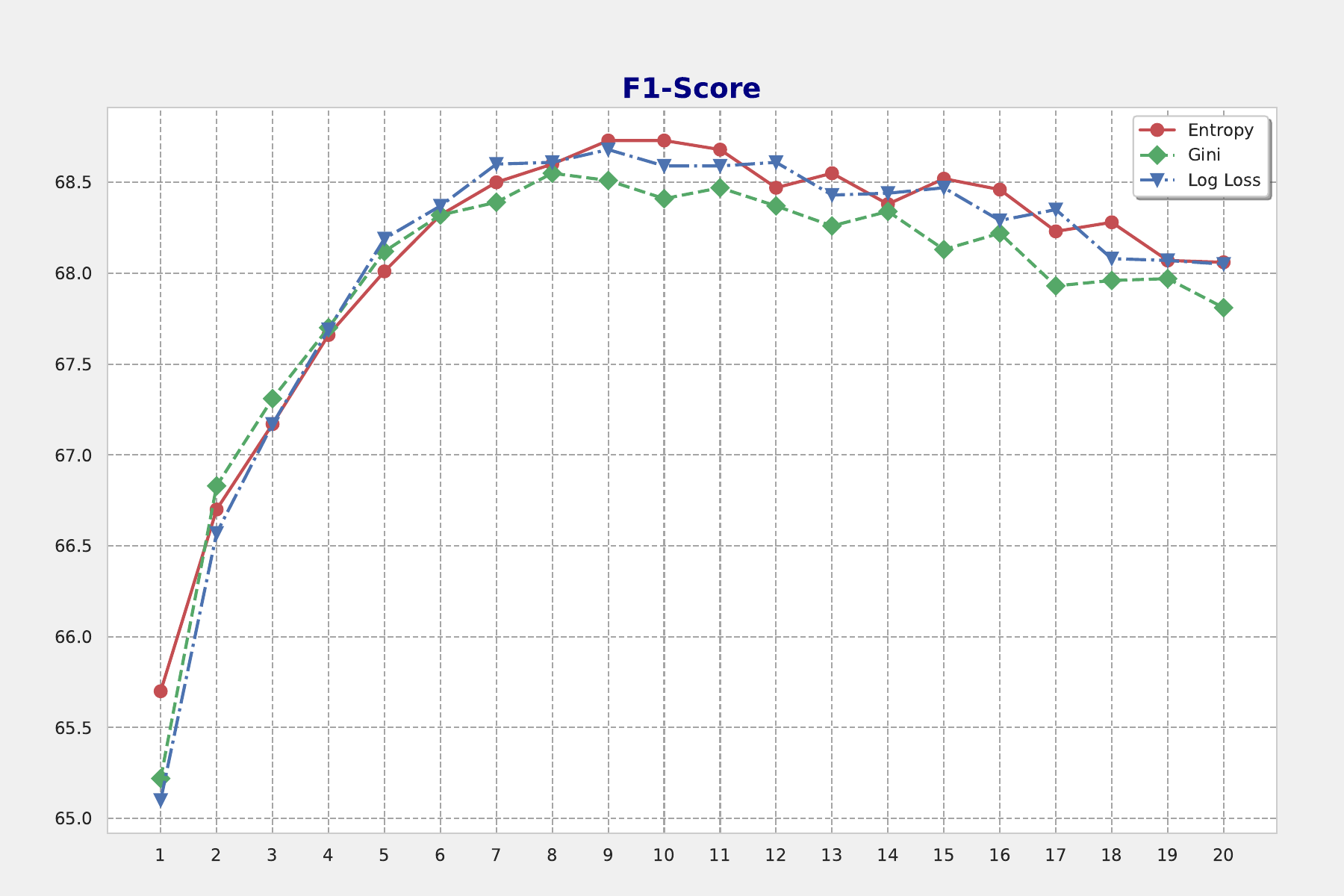}}
\subfloat[Precision]{\label{RFC:RF_Percision}\includegraphics[width=0.23\textwidth,height=4cm]{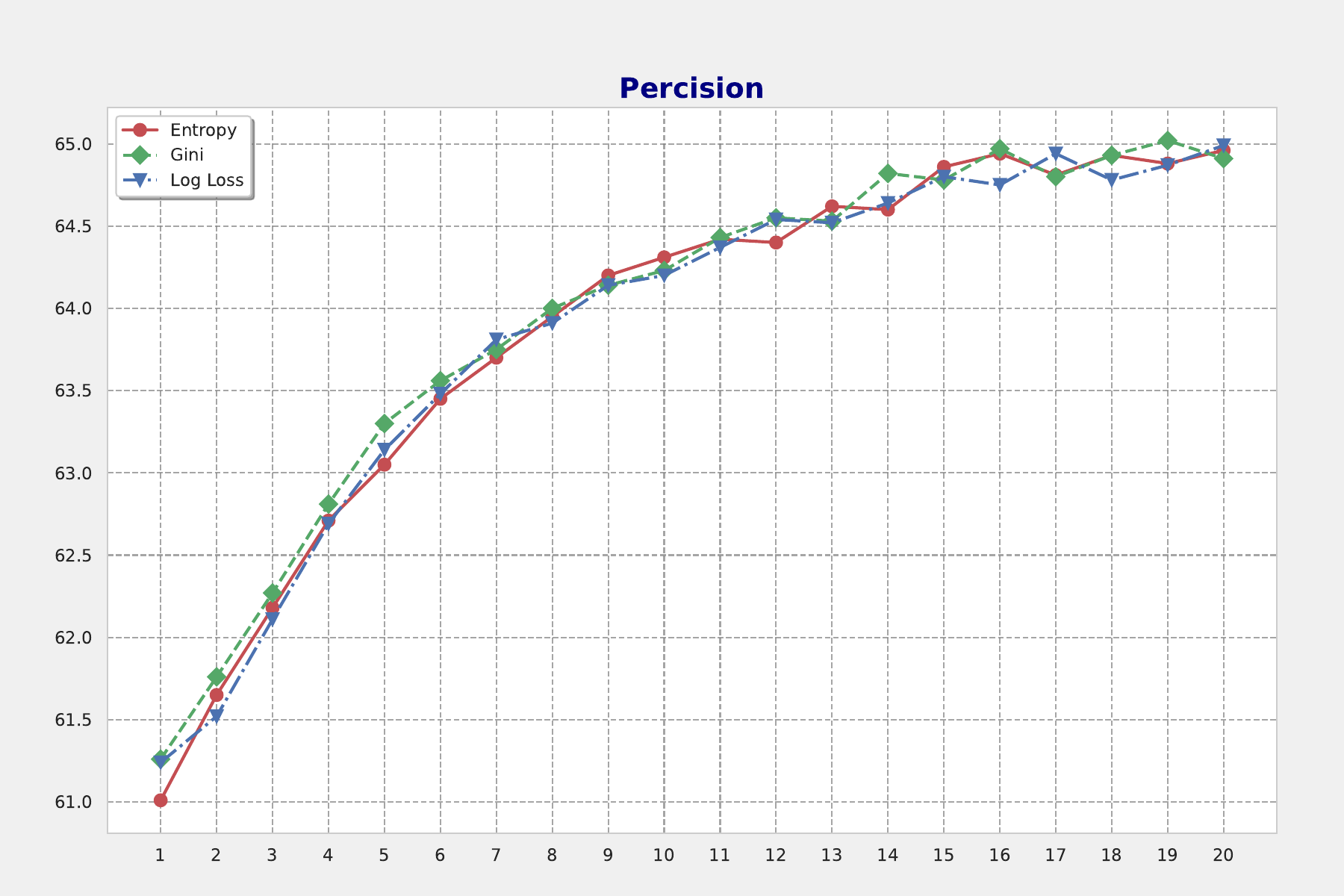}}\\
\subfloat[Recall]{\label{RFC:RF_Recall}\includegraphics[width=0.23\textwidth,height=4cm]{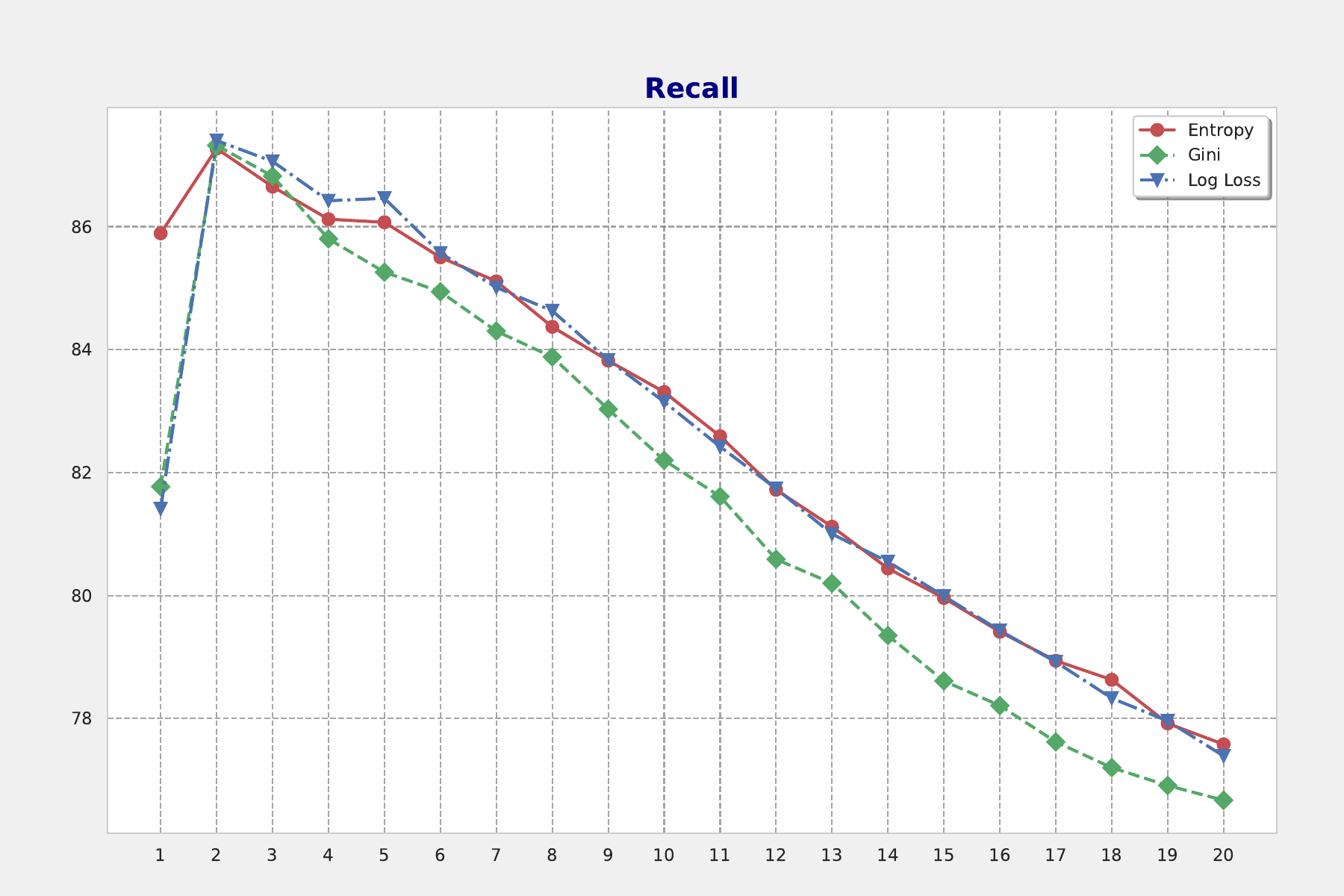}}
\subfloat[Jaccard Score]{\label{RFC:RF_JScore}\includegraphics[width=0.23\textwidth,height=4cm]{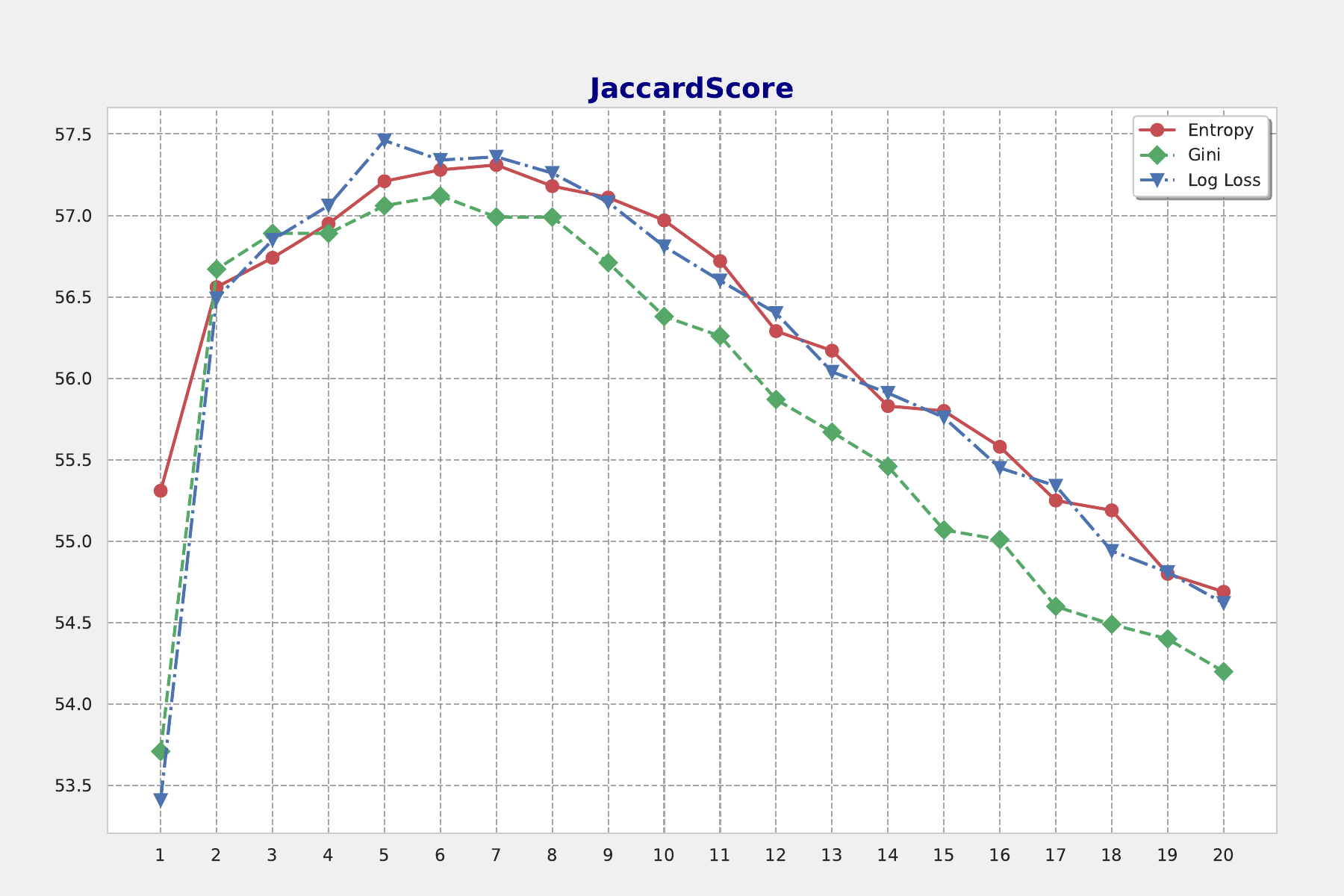}}\\
\caption{Random forests results on four metrics.} 
\label{fig:RF_results}
\end{figure}

Random forest combines multiple decision trees to reach a final result. 
Following the decision trees setting, for the random forest, we use three criteria: gini, entropy, and log loss with depths 1 to 20. Each depth is trained and tested 20 times. 
The four graphs in Figure \ref{fig:RF_results} depict the performance of a Random Forest classifier across different tree depths using F1-Score, Precision, Recall, and Jaccard Score as metrics.
\par
Figure \ref{RFC:RF_F1Score} presents the F1-Score, where it  rises with tree depth, peaking around depth 10 before slightly declining. 
It suggests a balance between precision and recall is best achieved at a moderate depth, with performance plateauing or slightly deteriorating at higher depths.
Figure \ref{RFC:RF_Percision} illustrates the precision. Similar to the F1-Score, precision improves as the tree depth increases.
Precision measures the accuracy of optimistic predictions, indicating that the model becomes better at correctly labeling positive instances up to a certain depth.
Between Entropy, Gini, and Log loss trend is identical till depth 9, but it changes after.

\par
Figure \ref{RFC:RF_Recall} presents the {Recall} results. Initially high, recall decreases as tree depth increases, reflecting the model's declining ability to identify all positive instances. 
It suggests that increasing the model's complexity (i.e., depth) might lead to overfitting, reducing its generalizability.
Figure \ref{RFC:RF_JScore} illustrates the Jaccard Score results. This score increases sharply between depths 2 and 7 and then decreases, mirroring the trend seen in Recall. The Jaccard Score, indicating the intersection over the union of the predicted and actual positive instances, suggests that the model's predictions are most aligned with the actual values at a moderate depth, with performance degrading as the model becomes too complex.
\par
Overall, the Random Forest model seems to perform optimally at a moderate tree depth across all metrics, with performance gains plateauing or reversing at higher depths, likely due to overfitting.
Table \ref{tab:RFClas} presents the final results where F1-Score, Precision, and Jaccard Score are best at Entropy while Recall is slightly better at Log loss. Mirroring the decision trees, Log loss is also the fastest in the Random Forest classifier. 
\begin{table}
	\centering
	\caption{Classification accuracy of random forest algorithm. RFC= Random Forest Criterion, D= Depth, F1=F1-score, Pre= Precision, Rec= Recall, JS= Jaccard Score, and T=Time}
	\label{tab:RFClas}       
	\begin{tabular}{lllllll}
		\hline\noalign{\smallskip}
		RFC & D & F1 & Pre & Rec & JS &  T \\
		\hline\noalign{\smallskip}
		Entropy &  9 &  \textbf{68.73}  & \textbf{64.20}  & 83.82  &\textbf{ 57.11} & 3.22\\
		Log Loss & 8 &  68.55   & 64.00  & \textbf{83.88}  & 56.99 & \textbf{2.19}\\
		GINI &  9 & {68.68}   & {64.14}  & {83.83}  & {57.08} & {3.26}\\
		\noalign{\smallskip}\hline
	\end{tabular}
\end{table}

\subsubsection{XGBoost}
An efficient and practical application of gradient boosting. It is a decision tree-based model where trees are added individually to the ensemble and fit the model. 
XGBoost has two types of boosters:  GbTree and Dart, both tree-based models. 
Both of their results are identical; hence, it does not have a plot as Figure \ref{fig:Dt_results} and Figure \ref{fig:RF_results}.
\begin{figure}[t]
	\centering
	\includegraphics[width=0.45\textwidth,height=6cm]{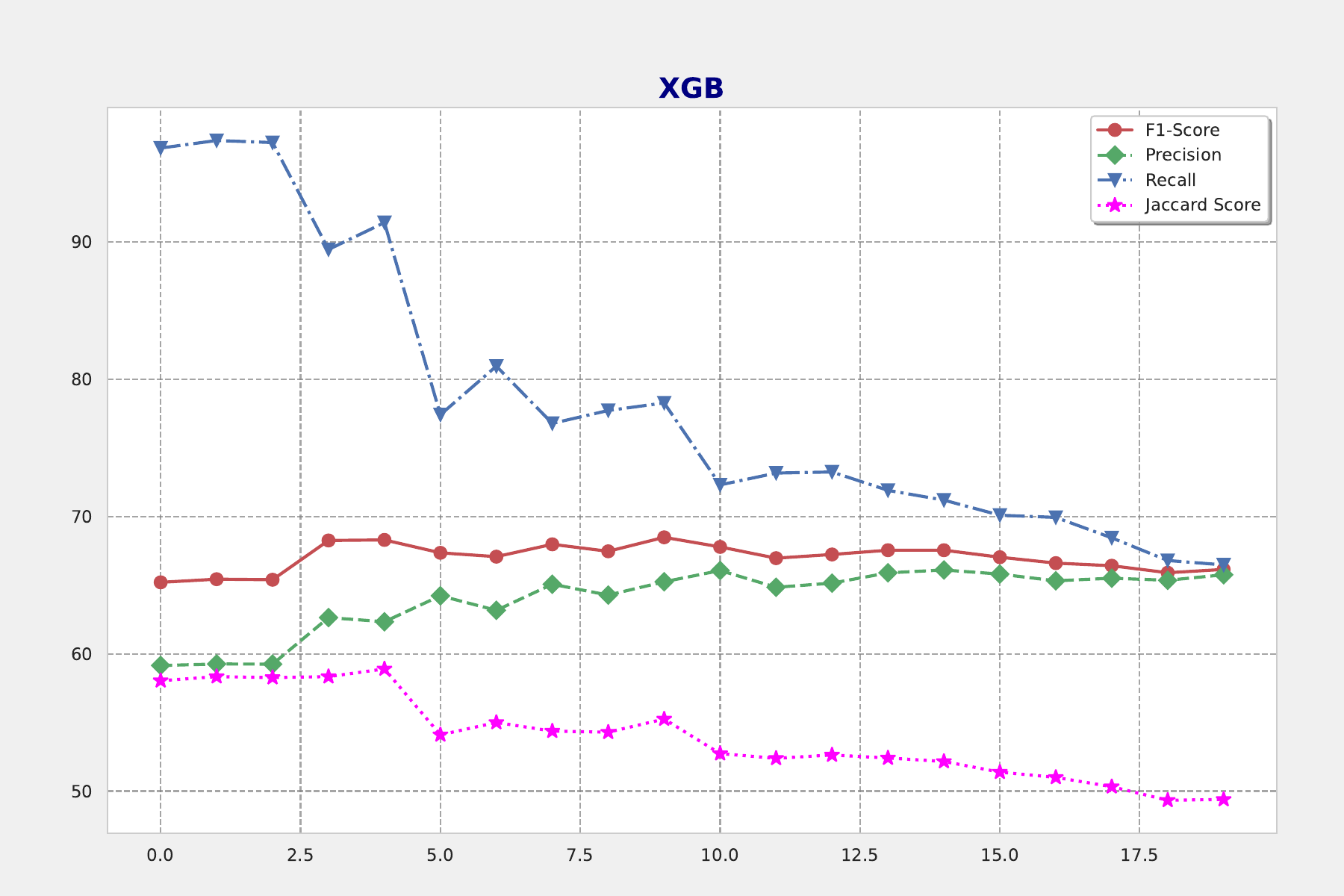}
	\caption{XGB results on four metrics.}
	\label{fig:XGB_results1}      
\end{figure}
\par
Figure \ref{fig:XGB_results1} presents the XGB results. 
F1-Score starts stable and then exhibits a slight increase with increasing depth but it fluctuates between high and lower results with depth and ends at lower results, suggesting that initially, the model benefits from more complex decision boundaries but might start to overfit as the depth increases.
Precision starts lower and increases with the depth with minor fluctuations. 
Recall bears an initial drop followed by a slight recovery but continues decreasing as depth increases. 
Jaccard Score is much lower than the other metrics and shows considerable volatility by starting higher and getting lower as tree depth increases. 
\par
The overall trend across all metrics indicates that the XGBoost model's ability to generalize does not improve with increasing depth beyond a certain point, with the most stable metric being Fq-Score and the most volatile being Precision. 
Table \ref{tab:XGBClas} presents the classification accuracy of XGBoost using these boosters while  the learning rate for all experiments is  $0.001$. 
The table illustrates that both tree-based boosters have the same results except for time; Gb Tree is faster than Dart. 
\begin{table}
	\centering
	\caption{Classification accuracy of XGBoost algorithm. XGC= XGBoost Criterion, D= Depth, F1=F1-score, Pre= Precision, Rec= Recall, JS= Jaccard Score, and T=Time}
	\label{tab:XGBClas}       
	\begin{tabular}{lllllll}
		\hline\noalign{\smallskip}
		XGC & D & F1 & Pre & Rec & JS &  T \\
		\hline\noalign{\smallskip}
		Dart &  10 &  68.49   & 65.25  & 78.28  & 55.25 & 75.44\\
		Gb Tree &  10 &  68.49   & 65.25  & 78.28  & 55.25 & 53.09\\
		\noalign{\smallskip}\hline
	\end{tabular}
\end{table}

\subsubsection{K-Nearest Neighbour}
\begin{figure}
	\centering
	\includegraphics[width=0.45\textwidth,height=6cm]{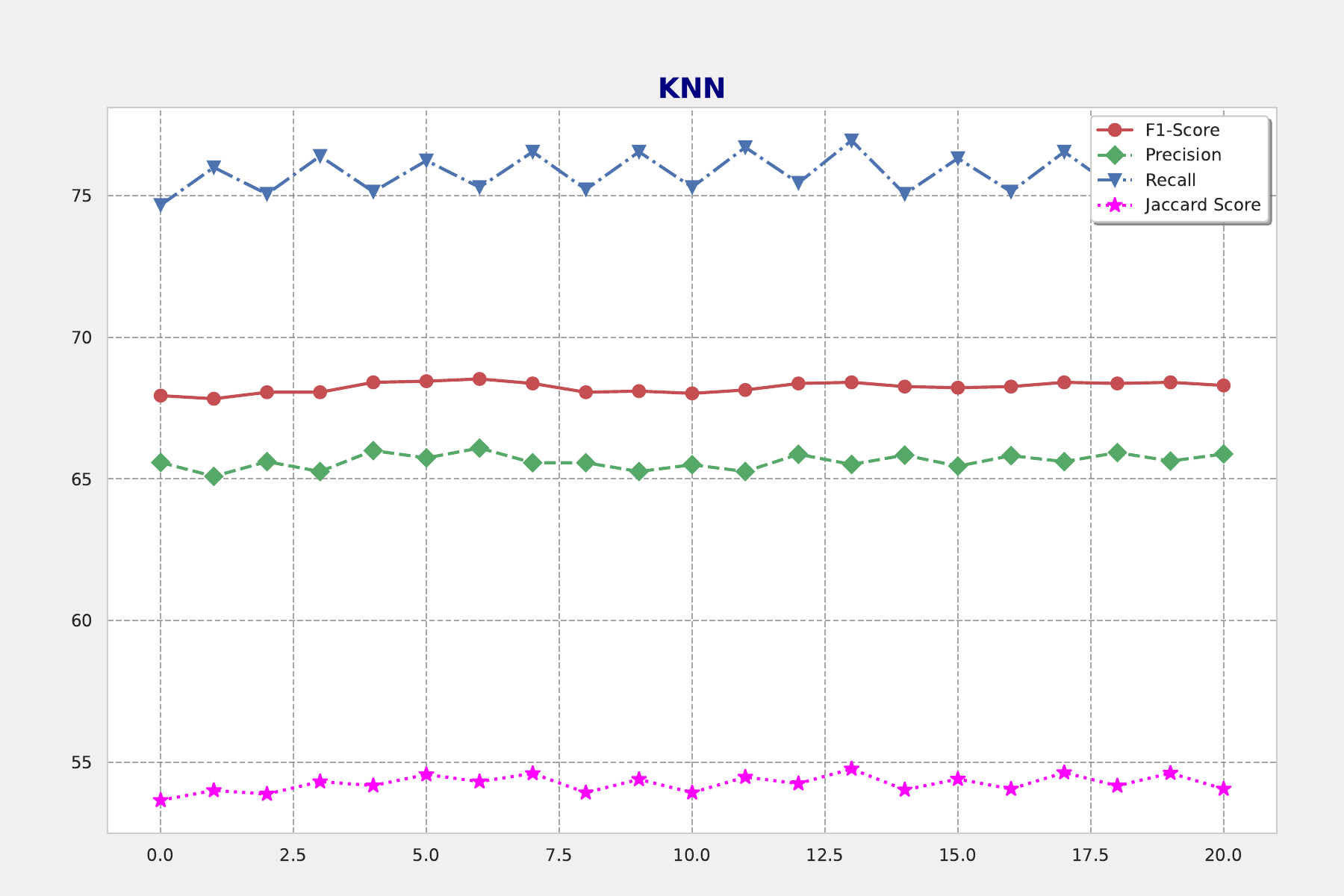}
	\caption{KNN results on all metrics.}
	\label{fig:All_Knn}      
\end{figure}

K-Nearest Neighbour (KNN) is a non-parametric model with good accuracy \cite{8862451}, where labeling decisions are made based on neighbors and the $K$ value indicates how many neighbors to choose. 
However, finding the optimal $K$ value is a hefty task. 
We perform the twenty experiments with different $K$ values to find the best $K$ and present the results in Figure \ref{fig:All_Knn}. 
\par
Figure \ref{fig:All_Knn} shows the performance of a KNN classifier on four metrics (F1-Score, Precision, Recall, and Jaccard Score) across twenty different values of $K$. 
The F1-Score remains relatively stable and high compared to other metrics, indicating a balanced classification. However, between different $K$ values, the F1-score is high on the odd value of $K$. 
Precision is stable across different $K$ values, showing consistent exactness in positive predictions. 
Recall is slightly lower than Precision but remains stable, suggesting a good rate of\textit{ true positive} identification. 
The Jaccard Score, however, is significantly lower than the other metrics across all $K$ values, indicating a lesser degree of similarity between the predicted and actual positive datasets. 
\par
The stability across different $K$ values suggests that the choice of K does not have a substantial impact on the classifier's ability to maintain its performance for this particular dataset and set of features, except the odd $K$ value yields better F1-Score than even. 
Table \ref{tab:KnnClas} concludes the overall results with the best $K$ value.
KNN performance is better on BiSND dataset compared to decision trees with better F1-Socre, Precision, and JS. 

\begin{table}
	\centering
	\caption{Classification accuracy of K-nearest neighbor algorithm. Algo=Algorithm, K= Number of neighbors, F1=F1-score, Pre= Precision, Rec= Recall, JS= Jaccard Score, and T=Time}
	\label{tab:KnnClas}       
	\begin{tabular}{lllllll}
		\hline\noalign{\smallskip}
		Algo & $K$ & F1 & Pre & Rec & JS &  T \\
		\hline\noalign{\smallskip}
		KNN &  17 &  68.57   & 68.11  & 73.77  & 54.83 & 1.53\\
		\noalign{\smallskip}\hline
	\end{tabular}
\end{table}
\begin{table*}[t]
	\centering
	\caption{Classification accuracy of all algorithms where finest values of each section are bold and highest results of the table are bold+italic. }
	\label{tab:classification}       
	\begin{tabular}{lllllll}
		\hline\noalign{\smallskip}
		Algorithm &Variant&  F1-score & Precision & Recall & Jaccard Score &  Time \\
		\noalign{\smallskip}\hline
		
		\multirow{1}{*}{DT} &Entropy &    {67.42}   & {63.21}  & {82.39}  &{55.69} & \textbf{\textit{0.19}}\\
		KNN&KNN &    68.57   & \textbf{68.11}  & 73.77  & 54.83 & 1.53\\
		\multirow{1}{*}{RF}&Entropy &  \textbf{68.73}  & {64.20}  & \textbf{83.82}  &\textbf{57.11} & 3.22\\
		\multirow{1}{*}{XGB}&Gb Tree &  68.49   & 65.25  & 78.28  & 55.25 & 53.09\\
		
		DNN&MLP & 66.83$\pm$0.46   & 64.25$\pm$0.96 & 71.18$\pm$3.87 & 50.91$\pm$1.42& 90.49\\
		\noalign{\smallskip}\hline\noalign{\smallskip}
		\multirow{3}{*}{GNN}&Only Nodes  &  66.92$\pm$0.46 &64.23$\pm$0.75 & 79.35$\pm$2.95 & 54.56$\pm$1.22 &  \textbf{25} \\
		&Un-directed  & 66.47$\pm$0.68   & {63.40}$\pm$0.93 & 81.19$\pm$2.57 & 54.82$\pm$1.25& 26 \\
		&Directed &  \textbf{67.25$\pm$0.41}  &\textbf{64.57$\pm$0.52} & \textbf{79.25$\pm$1.29} & \textbf{54.73$\pm$0.71}& {26}  \\
		\noalign{\smallskip}\hline\noalign{\smallskip}
		\multirow{5}{*}{GCL}&BGRL & 66.26$\pm$0.70 & 63.65$\pm$0.15 & 76.12$\pm$0.56 & 53.11$\pm$0.64& 1760 \\
		&GRACE &  67.39$\pm$0.37  & 63.27$\pm$0.15 & 77.75$\pm$0.31 & 53.20$\pm$0.13& 856 \\
		&DAENS\textsubscript{1D1D}  & 69.06$\pm$0.00   & 68.70$\pm$0.00  & 78.47$\pm$0.00  & 54.86$\pm$0.15 & 1376 \\
		&DAENS\textsubscript{1D2D} &69.08$\pm$0.06   & 69.49$\pm$0.54  & 78.07$\pm$0.22  & 55.80$\pm$0.21 & \textbf{665} \\
		&DAENS\textsubscript{2D2D} &\textbf{\textit{70.15$\pm$0.25}}   & \textbf{\textit{69.65$\pm$0.06}}  & \textbf{\textit{86.27$\pm$0.31}}  & \textbf{\textit{59.10$\pm$0.28}} & 715 \\

		\noalign{\smallskip}\hline
	\end{tabular}
\end{table*}
\subsubsection{ML Results Comparison}
\begin{figure}
	\centering
	\includegraphics[width=0.45\textwidth,height=6cm]{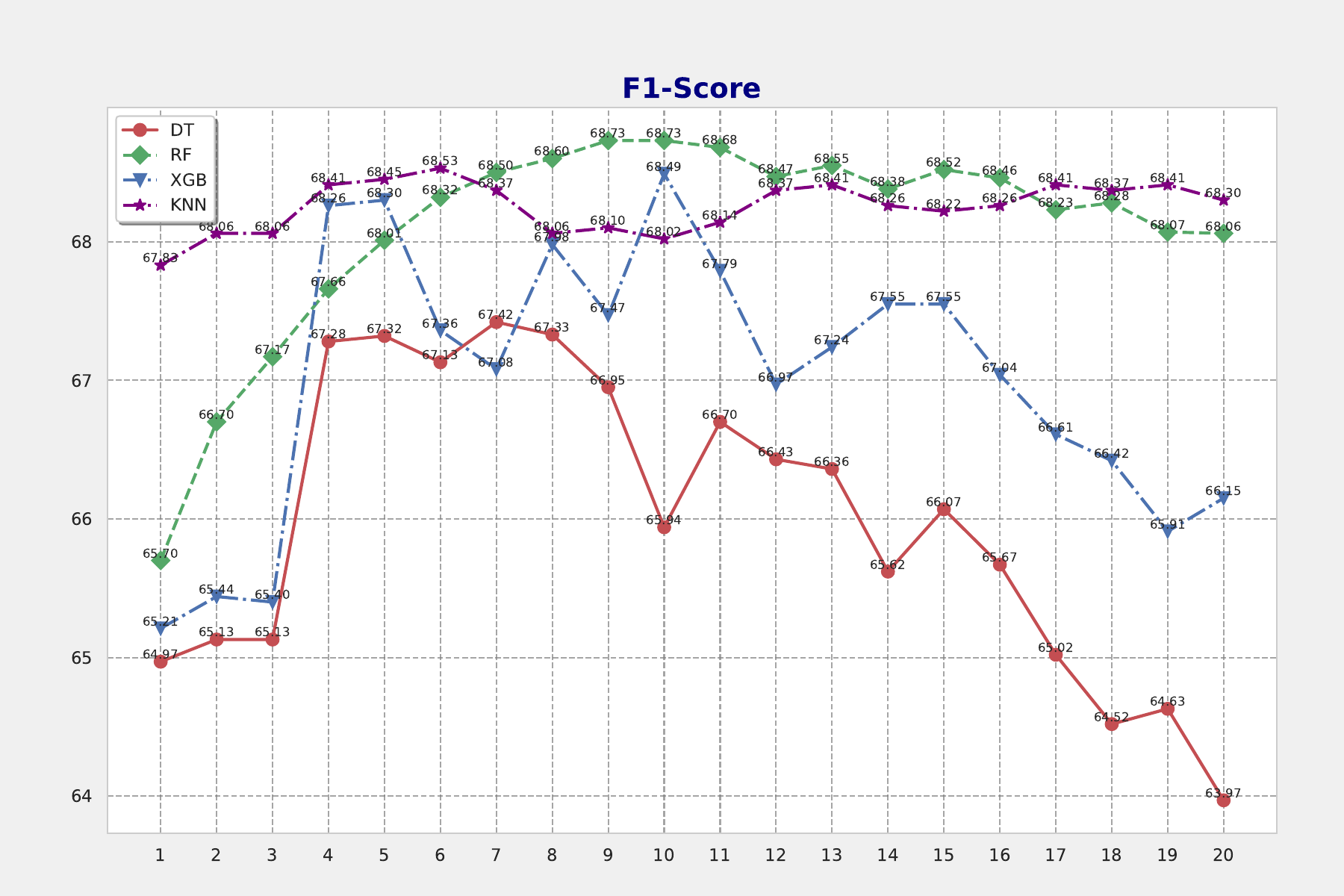}
	\caption{F1-Score accuracy of DT: Decision Trees, RF: Random Forest, XGB: XGBoost, and KNN: K-Nearest Neighbors.}
	\label{fig:Allf1}      
\end{figure}

Figure \ref{fig:Allf1}  compares the F1-Score accuracy of four different classifiers: Decision Trees (DT), Random Forest (RF), XGBoost (XGB), and K-Nearest Neighbors (KNN), across a range depth of the tree for DT, RF, and XGB, and the number of neighbors for KNN.
\textit{DT} shows considerable variability in F1-Score, with some depths performing significantly better or worse than others. This indicates that the model's performance is highly sensitive to the specific structure of the tree. However, It has the lowest results on average compared to other classifiers. 
\textit{RF} displays a more stable performance with less fluctuation in F1-Score. Its results increase with depth at a certain depth it starts decreasing, suggesting it is more robust to changes in the depth of the trees within the forest.
\textit{XGB} demonstrates less stability in F1-Score, starting with lesser accuracy then yields the best of its results, and again performance decreases with depth. 
\textit{KNN} has the most stable F1-Score, indicating that the choice of K does not substantially affect the performance or that the algorithm is not capturing the complexities of the dataset as effectively as the tree-based models.
In conclusion,  machine learning algorithms like Decision Tree (DT), K-Nearest Neighbors (KNN), Random Forest (RF), and XGBoost (XGB) show varying degrees of effectiveness. 
RF with an Entropy variant leads in F1-score (68.73\%) and JS (57.11\%), indicating a balanced trade-off between precision and recall. 
However, it lags in computational efficiency compared to DT. 
XGB demonstrates moderate performance across all metrics but requires significantly more computational time (53.09s).
This suggests that ensemble methods like RF and XGB are better at generalizing and dealing with overfitting than a single decision tree.
Table \ref{tab:classification} presents the best results of each machine learning classifier in the first four rows.

\subsection{Deep Neural Networks}
We use four layers of multi-layer perceptions (MLP) to perform these experiments. Relu as activation function and results are an average of 20 executions. 
Table \ref{tab:classification} presents that MLP performance is not better than ensemble machine learning on BiSND. 
Starting with the F1-score, MLP yields  just 0.13\% worse results than  KNN and 1.44\% worse than best performing Random Forest. 
The same trend follows with Precision where the difference is not as big to be noticed while Recall and JS tell a different story with the larger difference between MLP and ensemble machine learning. MLP execution time is also noticeably higher than ensemble machine learning. 
Table \ref{tab:classification} results conclude that BiSND is equally stable, robust, and important for ensemble machine learning and also for deep learning. 

\subsection{Graph Neural Networks}
This section performs experiments using graph neural networks (GNN) representation learning for classification as a downstream task on BiSND. 
There are three variants in Table \ref{tab:classification}: \textit{Only Nodes} refer to a graph without edges, \textit{Un-directed} refers to a graph with undirected edges, and \textit{Directed} refers to directional edges. 
The graphs' nature is depicted by three types of results: Directed, Undirected, and Only Nodes. 
Graph Neural Network (GNN) variants yield modest results. 
The Directed variant shows the best performance, slightly exceeding the others in precision, recall, and JS. 
However, the computational time (26s) is higher compared to the DT, KNN, and RF algorithms but lower than other methods. 
Overall GNN and DNN results are not as good as traditional machine learning. 
\par
Comparing these three variants, the directed graph has the best classification performance indicating that in BiSND, it is important if one node refers to another node with direction. 
Compared with ensemble machine learning and MLP, GNN results are not as good as ensemble machine learning while its results are better than DNN. 
GNN performance is better than DNN, it indicates that the relationship between nodes matters in BiSND and has a strong impact on classification performance. 
For the next section's results, we perform experiments with graph contrastive learning where we use the \textit{Directed} graph for further experiments due to its best performance in GNN.

\subsection{Graph Contrastive Learning}
Graph Contrastive Learning (GCL) is self-supervised learning where the model is trained without labels and classification is the downstream task. 
We use three SOTA methods for experiments: BGRL \cite{thakoor2021bootstrapped}, GRACE \cite{DeepGrace2020}, and DAENS \cite{DAENS} while there are three variants of DAENS.
Table \ref{tab:classification} presents that BGRL performance is not as good as ensemble machine learning while there are not much difference between machine learning and GCL results. GRACE results are identical to BGRL. 
\par
However, DAENS\textsubscript{1D1D}, DAENS\textsubscript{1D2D}, and DAENS\textsubscript{2D2D}, exhibit superior performance. 
BGRL and GRACE could not perform up to mark, and their results are almost equal to supervised learning and lesser than traditional machine learning. 
However, The variants of {DAENS} exhibit the finest overall performance, with the 2D2D variant achieving the highest F1-score (70.15\%), Precision (69.65\%), Recall (86.27\%), and JS (59.10\%) among the Table \ref{tab:classification}. 
This variant demonstrates a significant improvement over traditional methods and GNNs, albeit at a slightly higher computational cost.
The 1D2D variant of DAENS strikes a balance between performance and computational efficiency, featuring a notable F1-score of 69.08\% and the lowest run time (665s) among DAENS variants.
The 1D1D variant, while slightly less effective than the 2D2D variant, still surpasses other algorithms in terms of F1-score and Precision.

\par
Concluding the results of machine learning, deep learning, graph neural networks, and graph contrastive learning, Table \ref{tab:classification} yields that BiSND is a versatile and robust dataset that can generate stable performance among multiple machine learning domains.

\section{Conclusion}
\label{sec:Conclusion}
This study presents the \textbf{Bi}nary Classification \textbf{S}ocial \textbf{N}etwork \textbf{D}ataset (BiSND), a novel dataset designed for graph machine learning applications. The methodology for dataset creation is outlined for future researchers, with the dataset available in both tabular and graph formats.
We assessed the dataset's robustness using multiple machine learning methods, achieving F1-scores ranging from 67.66 (graph neural networks) to 70.4 (graph contrastive learning). These results suggest BiSND's suitability for classification tasks and potential for future enhancements.
Moreover, our findings indicate that tree-based methods excel with tabular data. In contrast, graph creation and self-supervised learning outperform it. This research provides a foundation for future studies in graph machine learning and social network analysis.

\bibliographystyle{IEEEtran}
\bibliography{biblo}


\end{document}